\documentclass[runningheads]{llncs}
\usepackage{array}
\usepackage{footnote}
\makesavenoteenv{tabular}
\usepackage{tabularray}
\usepackage{bm}
\usepackage{multirow}
\usepackage{amsmath}
\usepackage{amssymb}
\usepackage{pifont}
\usepackage{subcaption}
\usepackage{adjustbox}
\usepackage{listings}
\usepackage[T1]{fontenc}
\usepackage{graphicx}
\usepackage{hyperref}
\usepackage{color}  
\usepackage{xcolor}

\usepackage{epstopdf} 
\usepackage{caption}
\usepackage{float}
\hypersetup{
    colorlinks=true,       
    linkcolor=blue,        
    citecolor=blue,        
    filecolor=blue,        
    urlcolor=blue          
}

\lstdefinestyle{mystyle}{
    commentstyle=\color{green},
    keywordstyle=\bfseries\color{magenta}, 
    stringstyle=\color{red},
    basicstyle=\footnotesize\ttfamily,
    breakatwhitespace=false,
    breaklines=true, 
    captionpos=t,
    keepspaces=true,
    numbers=left, 
    numbersep=5pt,
    numberstyle=\tiny\color{gray},
    showspaces=false,
    showstringspaces=false,
    showtabs=false,
    tabsize=2,
    frame=tb,
    framerule=1pt,
    framextopmargin=0ex,
    framexbottommargin=0ex,
    framexleftmargin=0mm,
    language=Python,
    literate=
    {'}{{{\color{red}'}}}1,
    morekeywords={len, min, max}, 
    keywords=[2]{len, min, max},
    keywordstyle=[2]\color{blue} 
}
\begin{document}
%
%
\title{Low-Resource Crop Classification from Multi-Spectral Time Series Using Lossless Compressors}

\titlerunning{Low-Resource Crop Classification}
%
\author{
     Wei Cheng\textsuperscript{1}\orcidID{0000-0002-8190-5367} $^{\dagger}$
  \and Hongrui Ye\textsuperscript{1}\orcidID{0009-0006-0386-2732}$^{\dagger}$
  \and Xiao Wen\textsuperscript{1}\orcidID{0009-0008-9961-1376} $^{\dagger}$
  \and  Jiachen Zhang\textsuperscript{1}\orcidID{0009-0003-5090-0824}$^{\dagger}$
  \and Jiping Xu\textsuperscript{2}\orcidID{0009-0002-6659-9485}
  \and  Feifan Zhang\textsuperscript{1}\orcidID{0009-0008-4850-7472}$^{*}$
  }
\authorrunning{W. Cheng et al.}
%
\institute{{\textsuperscript{1}College of Information and Electrical Engineering,  China Agricultural University, Beijing 100083, China }\\{\textsuperscript{2}College of Science, China Agriculture University, Beijing 100083, China }\\
\textsuperscript{\dag}These authors contributed equally\\
\textsuperscript{*}Corresponding author\\
\email{\{weicheng, caucieeyhr, wenxiao, jiachen\_zhang, jipingxu, feifanzhang\}@cau.edu.cn}
}
\maketitle 

\begin{abstract}
Deep learning models have significantly improved the accuracy of crop classification using multispectral temporal data. However, they have complex structures with numerous parameters, requiring large amounts of data and costly training. In low-resource situations with fewer labeled samples or on low-computing devices, they perform poorly. Conversely, compressors are data-type agnostic, and non-parametric methods do not bring underlying assumptions. Inspired by this insight, we propose a non-training alternative to deep learning models, aiming to address these situations. Specifically, the symbolic representation module is proposed to convert the reflectivity into symbolic representations. The symbolic representations are then cross-transformed in both the channel and time dimensions to generate symbolic embeddings. Next, the Multi-scale Normalised Compression Distance (MNCD) is designed to measure the correlation between any two symbolic embeddings. Finally, based on the MNCDs, high quality crop classification can be achieved using only a \(k\)-nearest-neighbor classifier (\(k\)NN). The entire framework is ready-to-use and lightweight. Without any training, it outperforms, on average, 6 advanced deep learning models trained at scale on three benchmark datasets. It also outperforms more than half of these models in the few-shot setting with sparse crop labels.

\end{abstract}
\section{Introduction}

Multi-spectral temporal classification of crops plays a crucial role in growth monitoring, pest forecasting, crop yield estimation, and other agricultural applications. This helps enhance the efficiency and quality of agricultural production~\cite{alzhanov2024crop}. However, multi-spectral temporal data typically appears as a high-dimensional feature space with numerous features. Redundant information can restrict data fitting and representation capabilities~\cite{lin2007experiencing}.

Deep learning can automatically extract features from high-dimensional data, providing extensive opportunities~\cite{mavaie2023hybrid}. In the past decade, various neural network structures have been extensively researched, including networks based on multilayer perceptrons (MLP), convolutional neural networks (CNN), recurrent neural networks (RNN), and Transformer-based networks. Specifically, a method called Long-term Recurrent Convolutional Networks (LRCN) proposed by Donahue et al., which combines the strengths of CNN and RNN. By incorporating Long Short-Term Memory (LSTM) units, it is capable of establishing long-term temporal dependencies~\cite{donahue2015long}. Next, Jia et al. developed a spatiotemporal learning framework based on a dual-memory structure of LSTM. This framework further extends the performance of LSTM. It enables LSTM to establish temporal dependencies and spatial relationships between long-term and short-term events in time and space~\cite{jia2017incremental}. Furthermore, Xu et al. developed a deep learning method named DeepCropMapping (DCM) model. This model is based on the Long Short-Term Memory structure with an attention mechanism. The DCM model is trained on ARD time series data, allowing it to learn generalizable features during the training process~\cite{xu2020deepcropmapping}. Unfortunately, these models require significant computational resources and time, leading to challenges in processing long time series data. To address this challenge, Zhou and colleagues introduced the Informer model. It is based on transformer architecture and significantly improves the speed of long time series prediction~\cite{zhou2021informer}. Additionally, Nie et al. created the PatchTST model. The model significantly enhances the classification results of long-term prediction~\cite{nie2022time}. Meanwhile,Liu et al. introduced the iTransformer model, showcasing superior technical capabilities across various real-world datasets. The iTransformer model demonstrates high performance and strong generalization abilities,and it is highly effective for time series prediction~\cite{liu2023itransformer}. Currently, in the field of multivariate time series prediction, deep learning models such as CNNs, RNNs, and Transformers have all achieved highly advanced performance. However, in recent research, Zhang et al. introduced the LightTS architecture, which is a method that does not require any convolution or attention mechanisms. This study is the first to demonstrate that an MLP-based structure can also be highly efficient and accurate in multivariate time series prediction~\cite{zhang2022less}.

Although these backbone networks and their variants have achieved satisfactory classification accuracy, there are still shortcomings in the following aspects: (1) Deep learning models exhibit high complexity. They require a significant investment of resources during the training process and have a massive number of model parameters~\cite{zhang2022less}; (2) The cost of labeling for deep learning models is high. Obtaining the necessary data requires a significant investment in professional labor and resources. In the few-shot scenario with sparse labels, the performance of deep learning models is not satisfactory~\cite{xu2020deepcropmapping}.
Compressors are data-type agnostic, and non-parametric methods do not bring underlying assumptions~\cite{jiang2023low}. These features can effectively address the aforementioned problems. Inspired by this, the proposed Symbolic Representation Module is used to convert the reflectivity of all pixels into symbol representations. The symbolic representations are then cross-transformed in both the channel and temporal dimensions to generate symbolic embeddings. Next, the Multi-scale Normalised Compression Distance (MNCD) is designed to measure the correlation between any two symbolic embeddings. Finally, based on the MNCDs, classification is implemented using only a \(k\)NN. Our method revolves around simple lossless compressors. It offers advantages such as no trainable parameters and a lightweight structure, ensuring its wide practical application.

To sum up, contributions of this research are:
\begin{itemize}
    \item We treat a pixel as `text' and introduce compressor-based classification from text classification to multispectral temporal crop classification.
    \item A Symbolic Representation Module is proposed to convert the reflectivity of pixels into symbolic representations. Based on the symbolic representations, a method of cross-transformation in time and channel dimensions is then proposed to generate symbolic embeddings. The Multi-scale Normalised Compression Distance (MNCD) is then designed to measure the correlation between any two symbolic embeddings for subsequent classification.
    \item The entire framework is ready-to-use and lightweight. Without any training, it outperforms the average of 6 advanced deep learning models trained at scale on three benchmark datasets. It performs exceptionally well in few-shot scenarios, where the availability of labeled data is limited for effective neural network training. 
\end{itemize}

\section{Materials and Methods}
\subsection{Data Description}

Following the previous study~\cite{tarasiou2023vits}, we choose three benchmark datasets to evaluate the performance of all methods. The \textbf{German dataset}~\cite{DBLP:journals/corr/abs-1802-02080} covers a densely cultivated area of 102 \( \times \) 42 km\(^2\) north of Munich, Germany. It contains 17 different categories with 13 spectral bands. The \textbf{T31TFM-1618 dataset}~\cite{9854891} covers a densely cultivated S2 tile in France for the years 2016 to 2018 and includes 20 different categories with 13 bands. The \textbf{PASTIS dataset}~\cite{DBLP:journals/corr/abs-2107-07933} contains four different regions in France, covering over 4000 km$^2$ and including 18 crop categories with 33-61 acquisitions and 10 bands.

\begin{figure}[htbp] 
    \centering
    \includegraphics[width=\textwidth, keepaspectratio]{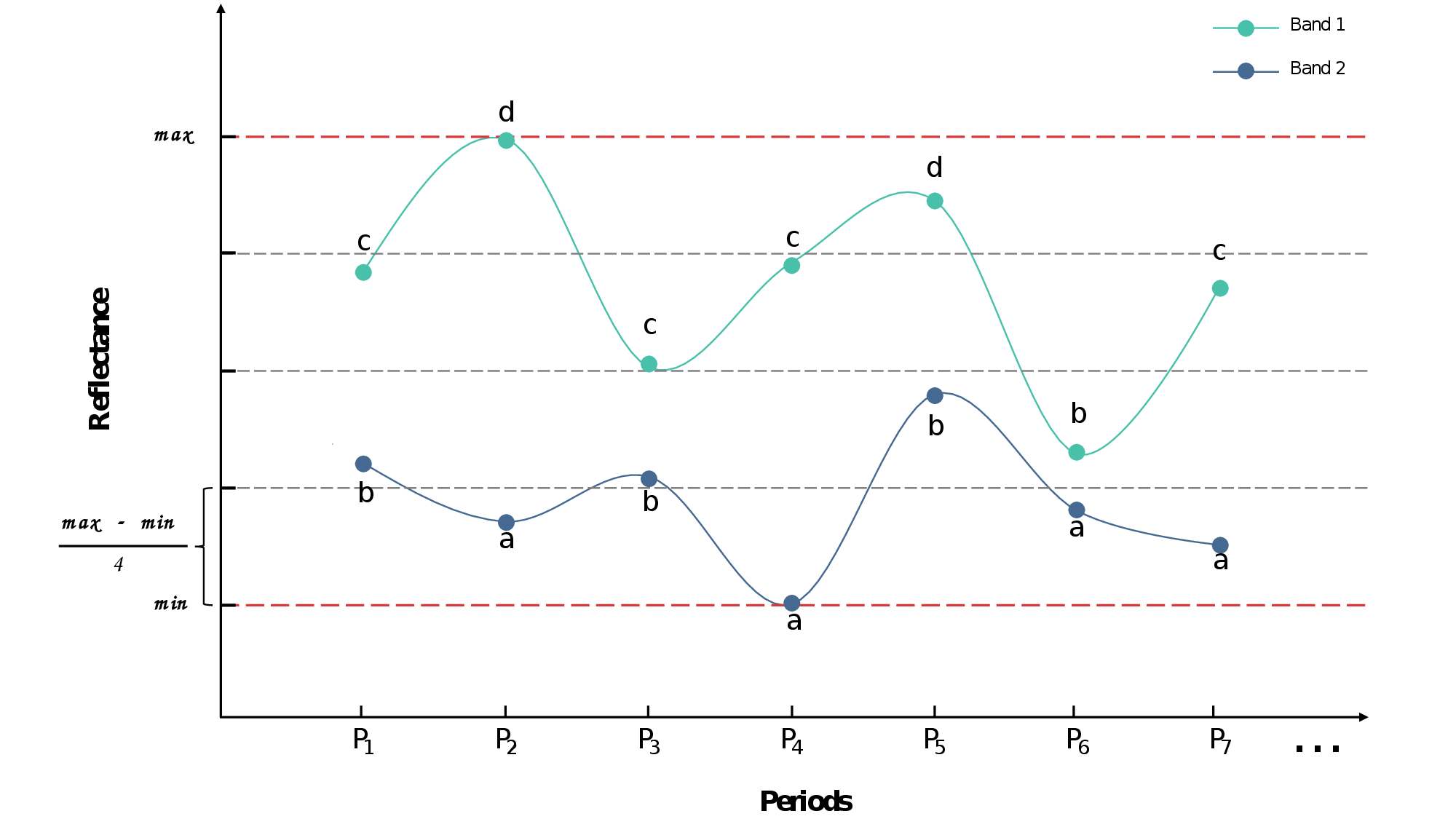} 
    \caption{These time series are discretized by using predetermined breakpoints $B$ to map the reflectivity to symbols. In the example above, with $l = 4$, the time series of band 1 is mapped to \textbf{cdcdcbc} and the time series of band 2 is mapped to \textbf{bababaa}.}
    \label{fig:symbolic_representation_algorithm}
\end{figure}

For each of the three datasets, we identify areas that are not heavily clouded 80\% of the time and exclude those that are. We also exclude background categories and those with less than 5 samples. In addition, due to the different resolutions of the different bands in the T31TFM-1618 dataset, we uniformly interpolate all bands to a maximum resolution of 48\( \times \)48 pixels. Following previous studies on temporal classification~\cite{wu2023timesnet}, 20\% of the pixel points from each crop type are randomly selected for the training dataset in the main experiment, while the remaining pixels are assigned to the test dataset. This leaves 20k training and 80k testing samples for the German dataset, 40k training and 170k testing samples for the T31TFM-1618 dataset, 15k training and 61k testing samples for the PASTIS dataset,  covering 266,36 ha land totally. To ensure the reproducibility of these experiments, we use a random seed of 32. Table~\ref{table:dataset} shows the number of samples and the training scale for each category in the main experiment.

\subsection{Design of Our Method}

\subsubsection{Symbolic Representation Module}

Lossless compressors are used in data storage and transmission scenarios. Their principle is to identify redundant information (such as repeated strings or symbols) in the data and replace it with a shorter representation, making them more suitable for data in symbolic form. Related research~\cite{lin2007experiencing} has also demonstrated the effectiveness of converting time series into symbolic representations. Inspired by this, we propose a simple and general module for symbolic representation.

\begin{figure}
\includegraphics[width=\textwidth]{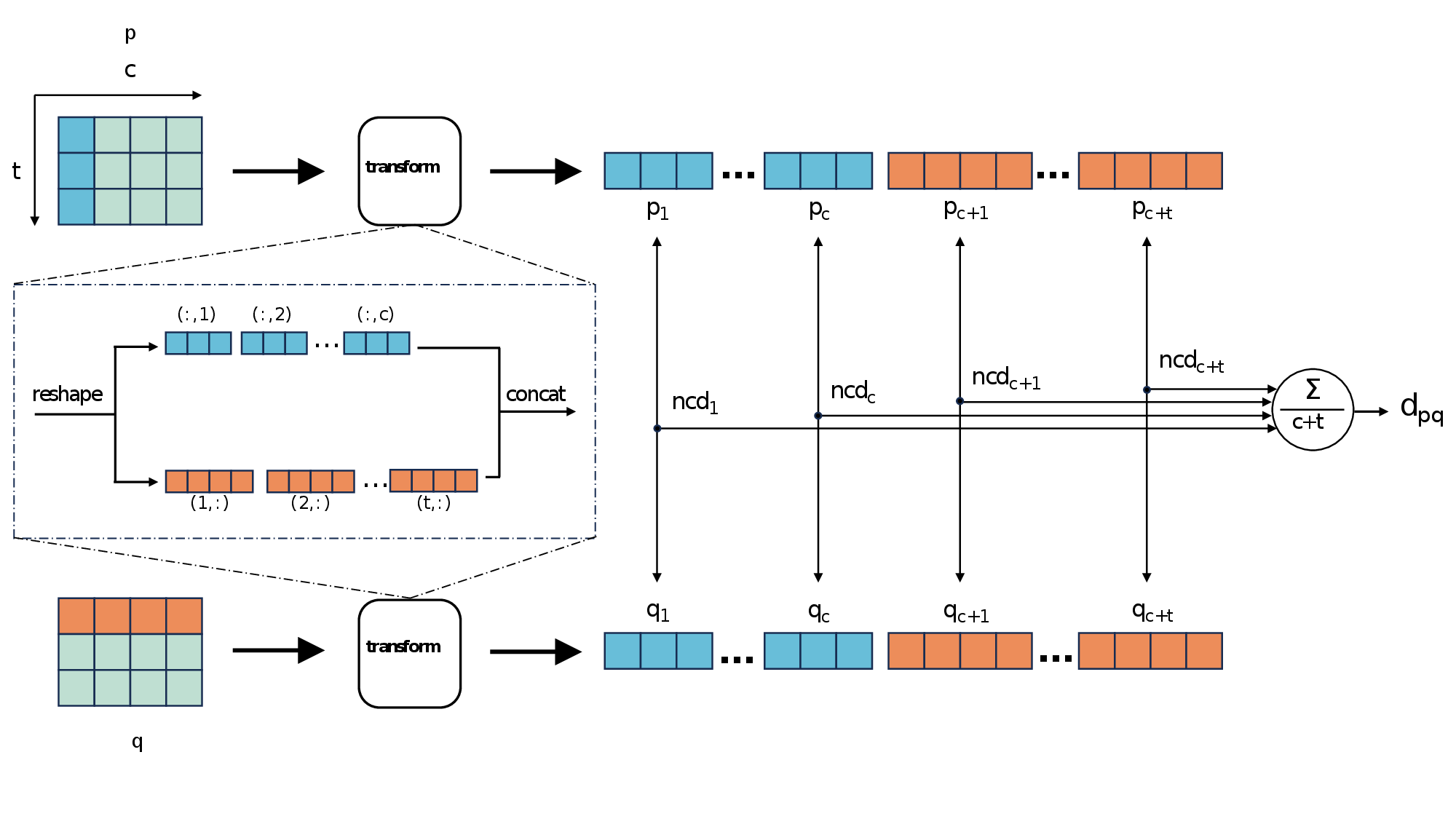}
\caption{An illustrative description of cross-transforming in the time and channel dimensions to obtain multi-scale NCD \(d_{p q}\) between pixels \(S_p\) and \(S_q\) after symbolic representation.}
\label{fig:Multiple NCD classification}
\end{figure}

Specifically, we first define an alphabet $\bm{\alpha}$ with up to 26 lowercase letters and 26 uppercase letters, where ${\alpha}_k$ represents the $k^{th}$ element of the alphabet, such as ${\alpha}_1={a}$, ${\alpha}_2={b}$, ..., ${\alpha}_{27}={A}$. At the same time, the maximum and minimum values of the reflectivity in all pixels are obtained, namely ${max}$ and ${min}$. Then, the reflectivity interval [${max}$, ${min}$] is divided into $l$ equal intervals to obtain breakpoints $\bm{B}$, as shown below:
\begin{equation}
\bm{B}=[\beta_1, \ldots, \beta_i, \ldots, \beta_{l+1}],
\end{equation}

where $l$ represents the length of the alphabet $\bm{\alpha}$, $\beta_1=min$, $\beta_{n+1}=max$, and $\beta_{i+1}-\beta_{i}=\frac{max-min}{l}$.

Once the breakpoints are determined, any reflectivity can be mapped to a symbolic representation. Given a pixel $\bm{x} \in \mathbb{R}^{t \times c}$ (where $t$ is the length of the time, and $c$ is the number of channels) in the original data, the symbolic representation value \(s_{ij}\) of any reflectivity \(x_{ij}\) can be obtained by the following mapping operation:

\begin{equation}
s_{ij}={\alpha}_k \text {, if } \beta_{k} \leq x_{ij}<\beta_{k+1}.
\end{equation}

For example, all reflectivities greater than or equal to the first breakpoint \(\beta_1\) and less than the second breakpoint \(\beta_2\) are mapped to the symbol \(a\); all reflectivities greater than or equal to the second breakpoint \(\beta_2\) and less than the third breakpoint \(\beta_3\) are mapped to the symbol \(b\). The rest can be inferred by analogy. Figure~\ref{fig:symbolic_representation_algorithm} shows the details of this module. This mapping converts the reflectivity of all pixels into symbolic representations that are more suitable for compressors.

\subsubsection{Compressor-Based Crop Classification}

Previous work combines band data at different time points for a single pixel to obtain information about the pixel~\cite{xu2020deepcropmapping}. Our method not only continues to use this time-series band combination, but also introduces a new dimension: combining data from the same pixel in different bands in chronological order. Then, by combining these two data sequences, we can more comprehensively describe the characteristics of a single pixel. Specifically, given a pixel after symbolic representation $\bm{s} \in \mathbb{R}^{t \times c}$, the following modeling is given:

\begin{equation}
\label{Calculate the average NCD}
\bm{S}=f(\bm{s})=\operatorname{concat}\left(\boldsymbol{s}^1, \ldots, \boldsymbol{s}^c, \boldsymbol{s}^{c+1}, \ldots, \boldsymbol{s}^{c+t}\right),
\end{equation}

where the function $f(\cdot)$ is the cross-transformation method of $\bm{s}$. The function \(\text{concat}(\cdot)\) denotes the concatenation operation of the symbolic representation. $\bm{S}$ stands for the symbolic embedding of $\bm{s}$. For any $\bm{s}^i$ it holds that:

\begin{equation}
\boldsymbol{s}^i = \begin{cases}
{\left[s_{1i}, s_{2i}, \ldots, s_{ti}\right]}, & \text{ if } 1 \leq i \leq c \\
{\left[s_{(i-c)1}, s_{(i-c)2}, \ldots, s_{(i-c)c}\right]}, & \text{ if } c < i \leq c+t.
\end{cases}
\end{equation}

The above cross-transformation can be easily achieved using methods such as \texttt{numpy.reshape()} and \texttt{numpy.concatenate()}, which are based on numerical computing libraries such as NumPy.

For any two pixels $\bm{x}_p$ and $\bm{x}_q$, the regularity between them can be measured by the correlation between their corresponding symbolic embeddings, $\bm{S}_p$ and $\bm{S}_q$. And the regularity between two symbolic embeddings can be measured by the Nomalised Compression Distance (NCD)~\cite{li2004similarity} of lossless compressors. Lossless compressors optimise the representation of information by assigning shorter codes to more frequent symbols. The idea is that: (1) compressors are good at capturing regularities; (2) objects within the same category exhibit more regularities than those from different categories~\cite{jiang2023low}.

Quantifying this regularity as $d_{p q}$, its definition is as follows:

where the function $g(\cdot)$ represents the operation of Multi-scale NCD (MNCD) for $\bm{S}_p$ and $\bm{S}_q$, and $d_{pq}$ represents the MNCD between $\bm{S}_p$ and $\bm{S}_q$. The definition of NCD is as follows:


\begin{equation}
N C D(m, n)=\frac{C(m || n)-\min \{C(m), C(n)\}}{\max \{C(m), C(n)\}}
\end{equation}

\begin{equation}
d_{p q}=g(\bm{S}_p, \bm{S}_q)=\frac{1}{c+t} \sum_{k=1}^{c+t} N C D\left(\bm{s}_p^k, \bm{s}_q^k\right),
\end{equation}

$C(m)$ denotes the length of $m$ after compression using a lossless compressor. $C(m || n)$ refers to the compressed length of \(m\) and \(n\) after being concatenated. This method can compute the MNCD between any symbolic embedding in the training set and symbolic embedding in the test set, and then construct a MNCD matrix. Finally, $k$NN is used to classify each symbolic embedding in the test set.

It is worth noting that when using a \(k\)NN, if \(k > 1\), there may be situations where the number of samples from several different categories is the same among the \(k\) closest samples. To break the tie, we suggest finding the training set sample with the smallest NCD from the test sample among these categories. Then, the category of this training set sample will be used as the predicted category for the test sample. The code for the entire method is presented as Algorithm~\ref{alg:compress}.

\begin{lstlisting}[language=Python, basicstyle=\small\ttfamily, caption={Python Code for Compressor-Based Crop Mapping}, label=alg:compress]
import gzip
import numpy as np

for (s1 , _ ) in test_set:
    Cs1 = len(gzip.compress(s1.encode()))
    distance_from_s1 = []
    for ( s2 , _ ) in train_set:
        Cs2 = len(gzip.compress(s2.encode()))
        s1s2 = ''.join([s1, s2])
        Cs1s2 = len(gzip.compress(s1s2.encode()))
        ncd = (Cs1s2 - min(Cs1, Cs2)) / max(Cs1, Cs2)
        distance_from_s1.append(ncd)
    sorted_idx = np.argsort(np.array(distance_from_s1))
    top_k_labels = [train_set[i][-1] for i in sorted_idx[:k]]

    # break ties
    unique_labels, label_counts = np.unique(top_k_labels, return_counts=True)
    most_common_labels = unique_labels[label_counts == np.max(label_counts)]
    if len(most_common_labels) == 1:
        s1_predicted_label = most_common_labels[0]
    else:
        indices = np.where(np.isin(top_k_labels, most_common_labels))[0]
        s1_predicted_label = top_k_labels[indices[0]]
\end{lstlisting}

\subsection{Implementations Details}
The proposed framework uses \textit{gzip} as the compressor. Under the premise of symbolic representation, the alphabet lengths \(l\) for the three datasets are all set to 22. The \(k\) value of \(k\)NN is set to 2. These parameters will be discussed in the subsequent Analyses section.

\subsection{Baselines}
Several representative deep learning models are selected for comparative experiments. They are mainly divided into four groups: MLP-based, RNN-based, CNN-based and Transformer-based. We select both classic baselines and state-of-the-art (SOTA) models. The MLP-based category includes MLP and DLinear~\cite{zeng2023transformers}. The RNN-based category includes Long Short-Term Memory (LSTM)~\cite{Hochreiter1997LSTM} and DeepCropMapping (DCM)~\cite{xu2020deepcropmapping}. The CNN-based group includes TempCNN~\cite{wu2020temp} and TimesNet~\cite{wu2023timesnet}. The Transformer-based category includes Transformer~\cite{vaswani2017attention} and Informer~\cite{zhou2021informer}. The parameters specific to these models are as follows and all models have converged in both the complete dataset and the few-shot settings:

For the MLP, we configure it with a single hidden layer of 128 neurons, a batch size of 32, and an initial learning rate of 0.001. We implement early stopping with a patience of 10 iterations and a validation fraction of 0.1, along with a maximum of 100 training epochs. The same configuration is used for the MLP in the few-shot setting.

For the DLinear, DCM, TimesNet and all Transformer-based models, we follow the optimal hyper-parameters recommended in the previous studies and corresponding repositories. Specifically, the batch size is set uniformly to 36, the dropout rate to 0.5, the number of epochs to 100 and the patience for early stopping is set to 10. In terms of model architecture, DLinear consists of three encoder layers and one decoder layer, each featuring a 256-dimensional feed-forward network. DCM, an attention-based bidirectional LSTM model, includes 2 LSTM layers with a hidden size of 256. The Transformer model is supported by three encoder layers and one decoder layer, also emphasizing a 256-dimensional feed-forward network. Similarly, Informer includes three encoder layers and one decoder layer. The same configuration is used for these models in the few-shot setting.

For the LSTM and TempCNN models, we follow the optimal hyperparameters recommended in the paper by Rubwurm et al.~\cite{rubwurm2019breizhcrops} and its corresponding repository. Specifically, the bidirectional LSTM stacks 4 layers with 128 hidden units. The learning rate is set to \(9.88 \cdot 10^{-3}\) with a decay rate of \(5.26 \cdot 10^{-7}\), and the number of epochs is 17. TempCNN uses a kernel size of 7 with 128 hidden units, a learning rate of \(2.38 \cdot 10^{-4}\), a decay rate of \(5.18 \cdot 10^{-5}\) and 11 epochs. For the few-shot setting, the number of epochs for the three models has been increased to 200 and 100.

\subsection{Evaluation Metric}
We evaluate the classification performance of each model based on two commonly-used indices, i.e., overall accuracy (OA), Mean Intersection over Union (MIoU). Specifically, the OA is the percentage of all correctly predicted instances in the dataset. The MIoU is the average ratio of overlap to union for predicted and actual segments
\section{Results}
\subsection{Comparison with Deep Learning Models}

The quantitative classification results in terms of OA and mIoU are shown in Table~\ref{table:results} for three datasets. Overall, the classification performance of the proposed method is significant. In terms of OA, it outperforms 5, 8 and 5 deep learning models on three datasets respectively. In particular, it outperforms all MLP-based and RNN-based models, demonstrating excellent classification performance. On the German dataset, however, it is not only 2.14\% lower on average than the CNN-based model, but also 1.30\% lower than the Informer. The performance gap is similar on the PASTIS dataset. A possible reason for this is that both CNN-based and Transformer-based models are able to learn high-level representations over time~\cite{wu2023timesnet}, which is crucial for classification tasks. However, such a result is acceptable given that our method can compete with massively trained deep learning models without any training. This is not always the case on the T31TFM-1618 dataset. The classification performance of our method outperforms all deep learning models. Compared to the CNN-based models and Transformer-based models, it averages 1.34\% and 0.51\% higher on OA, demonstrating excellent performance.

\begin{table}[h]
\centering
\caption{Classification results obtained with our method and baseline models. \textcolor{red}{Red} highlights the models that outperform our method.}
\label{table:results}
\begin{adjustbox}{max width=\textwidth}
\begin{tblr}{
  cells = {c},
  cell{1}{1} = {c=2,r=2}{},
  cell{1}{3} = {r=2}{},
  cell{1}{4} = {c=3}{},
  cell{1}{7} = {c=3}{},
  cell{1}{10} = {c=3}{},
  cell{3}{1} = {r=2}{},
  cell{5}{1} = {r=2}{},
  cell{7}{1} = {r=2}{},
  cell{7}{4} = {fg=red},
  cell{7}{5} = {fg=red},
  cell{7}{6} = {fg=red},
  cell{7}{10} = {fg=red},
  cell{7}{11} = {fg=red},
  cell{7}{12} = {fg=red},
  cell{8}{4} = {fg=red},
  cell{8}{5} = {fg=red},
  cell{8}{6} = {fg=red},
  cell{8}{9} = {fg=red},
  cell{8}{10} = {fg=red},
  cell{8}{12} = {fg=red},
  cell{9}{1} = {r=2}{},
  cell{9}{8} = {fg=red},
  cell{9}{11} = {fg=red},
  cell{10}{4} = {fg=red},
  cell{10}{5} = {fg=red},
  cell{10}{6} = {fg=red},
  cell{10}{8} = {fg=red},
  cell{10}{10} = {fg=red},
  cell{10}{11} = {fg=red},
  cell{10}{12} = {fg=red},
  cell{11}{1} = {c=2}{},
  hline{1,12} = {-}{0.3em},
  hline{3, 5, 7, 9, 11} = {-}{0.1em},
  vline{2, 3, 4, 7, 10} = {-}{0.1em},
  hline{2} = {-}{0.1em}
}
model       &             & trainable parameter & German &       &       & T31TFM-1618 &       &       & PASTIS &       &       \\
            &             &                     & OA     & AA    & mIoU  & OA          & AA    & mIoU  & OA     & AA    & mIoU  \\
MLP         & MLP         & \ding{51}                   & 69.62  & 52.00 & 34.24 & 68.49       & 44.37 & 31.11 & 92.06  & 67.68 & 60.40 \\
            & DLinear     & \ding{51}                   & 68.41  & 84.38 & 40.34 & 77.18       & 70.54 & 63.36 & 84.58  & 93.19 & 64.08 \\
RNN         & LSTM        & \ding{51}                   & 97.41  & 96.41 & 91.59 & 90.99       & 81.33 & 75.69 & 97.37  & 96.08 & 92.03 \\
            & DCM         & \ding{51}                   & 97.57  & 96.32 & 94.06 & 81.91       & 79.00 & 72.13 & 98.29  & 97.05 & 95.12 \\
CNN         & TempCNN     & \ding{51}                   & 98.61  & 97.79 & 96.18 & 95.74       & 92.26 & 88.66 & 98.67  & 98.23 & 96.47 \\
            & TimesNet    & \ding{51}                   & 98.75  & 98.19 & 96.56 & 96.68       & 95.40 & 92.87 & 99.08  & 97.26 & 97.03 \\
Transformer & Transformer & \ding{51}                   & 97.47  & 96.37 & 92.66 & 96.71       & 95.73 & 91.23 & 98.32  & 98.20 & 94.93 \\
            & Informer    & \ding{51}                   & 98.91  & 98.40 & 96.82 & 96.54       & 95.68 & 91.10 & 98.92  & 98.38 & 96.81 \\
Ours        &             & \ding{55}                   & 97.61  & 96.89 & 94.10 & 96.88       & 95.42 & 91.43 & 98.37  & 97.64 & 95.39 
\end{tblr}
\end{adjustbox}
\end{table}

\begin{figure}[h]
\centering
\includegraphics[width=\textwidth]{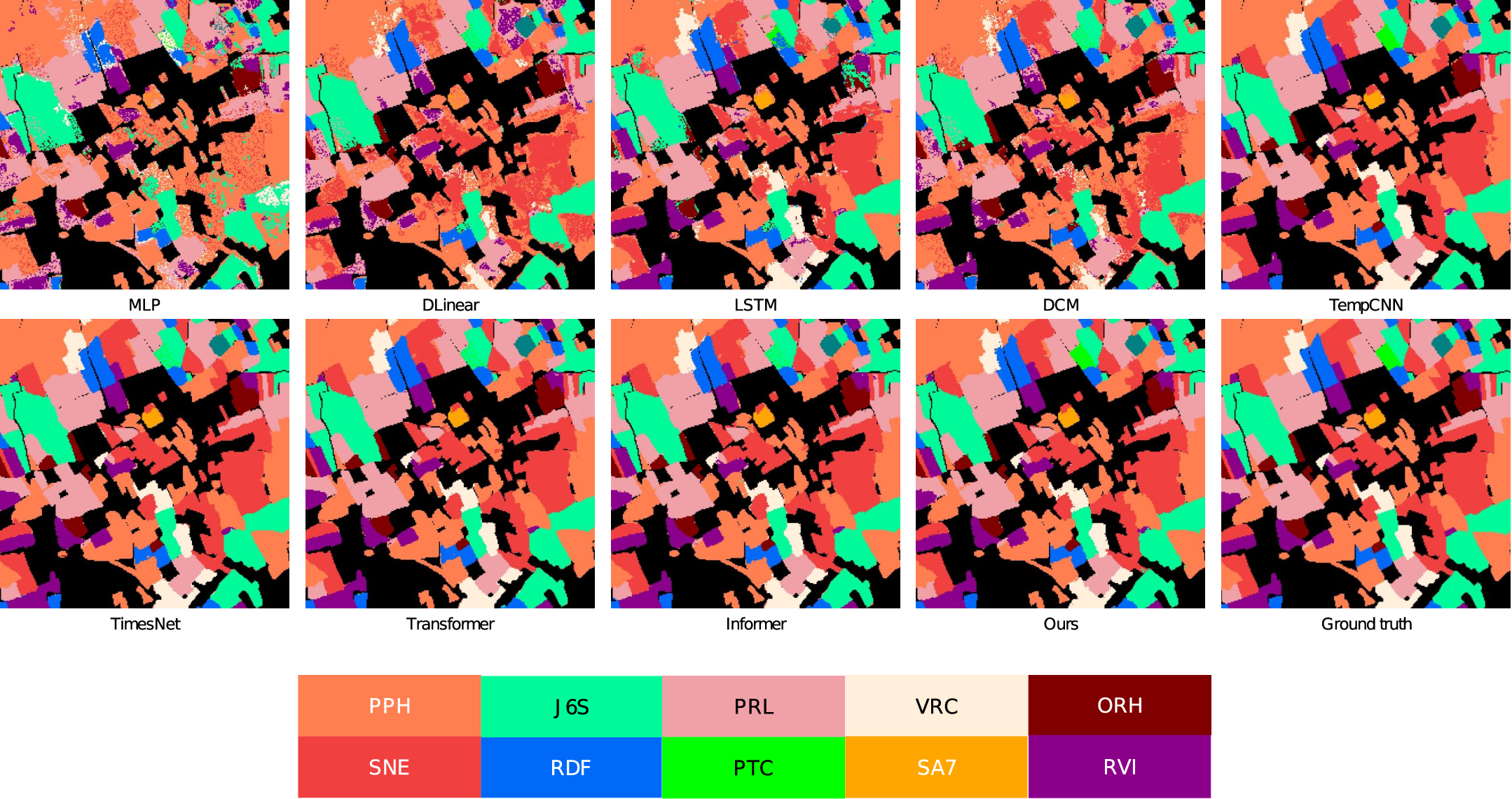}
\caption{Classification maps obtained from different models on a typical plot.}
\label{fig:result_visual}
\end{figure}

We also select a representative plot in Figure~\ref{fig:result_visual} with many boundaries and small fragmented fields, which thoroughly tests the classification capabilities of models. It is evident that the first four models easily make a lot of misjudgements, resulting in significant salt-and-pepper noise. In addition, the classification performance at the boundaries is rather coarse. In contrast, CNN-based models, Transformer-based models and our method show significant advantages. The boundaries are smoother and more refined, with almost no large-scale misclassifications. Considering that our method achieves classification performance comparable to deep learning models without requiring any training, it demonstrates significant advantages and practical value.

\subsection{Few-Shot Learning}

\begin{figure}[h]
\centering
\includegraphics[width=\textwidth]{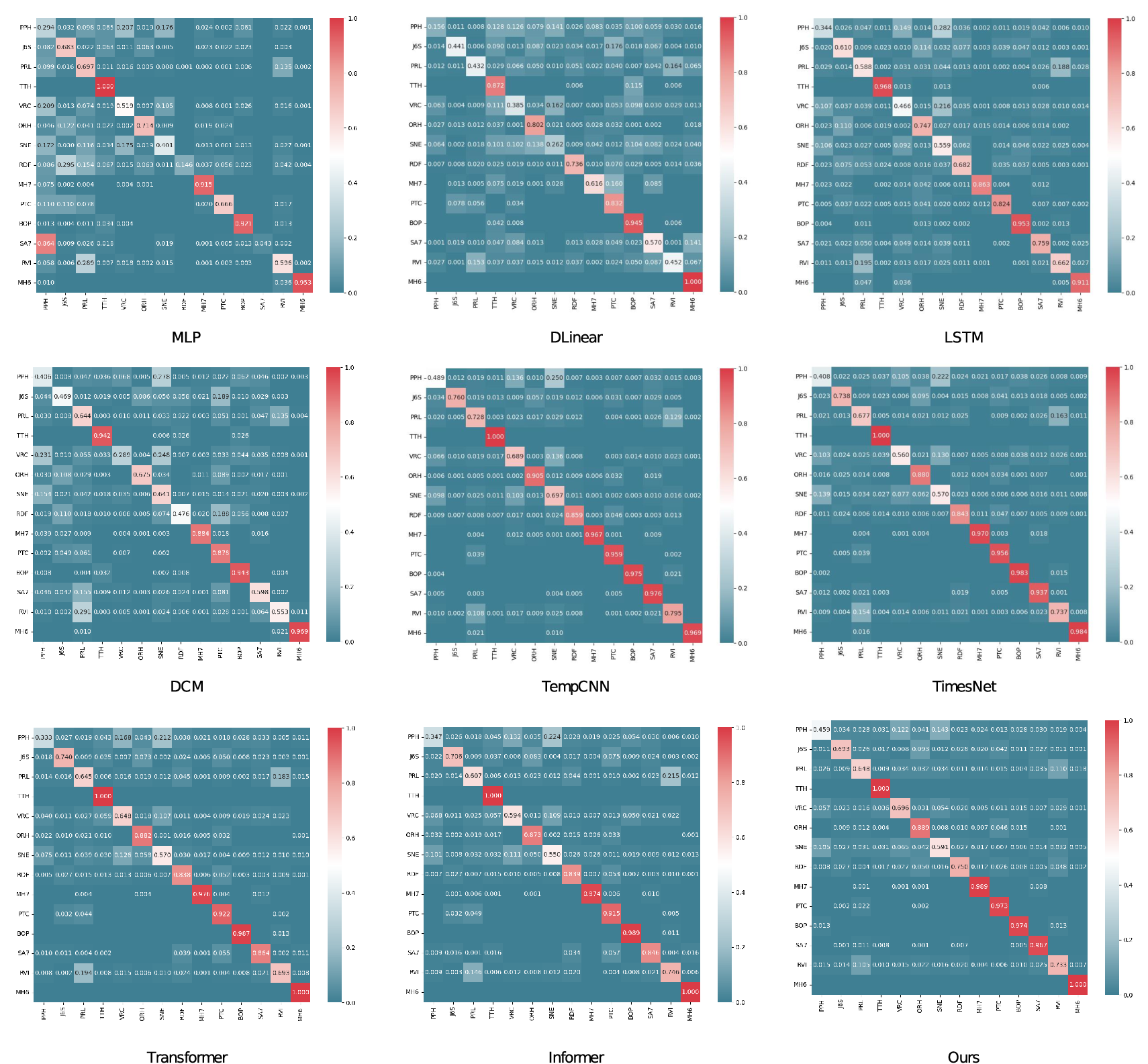}
\caption{Confusion matrix for 50-shot crop classification results on the T31TFM-1618 dataset.}
\label{fig:confusionM}
\end{figure}

We also compare our method with deep learning models in the few-shot setting across three datasets. Following the method of Jiang et al.~\cite{jiang2023lowresource}, we use \( n \)-shot labeled examples per category from the training dataset, where \( n \) sets to 50, 20, 10, 5. To ensure the robustness of our results, we run the experiment five times, using a different random seed to select the subset for each run. We then calculate the mean and 95\% confidence intervals over the five trials. The five random seeds used are 2024, 21, 32, 400 and 47. The detailed results are listed in the Table~\ref{table: fewshot}.

\begin{table}[h]
\centering
\caption{Results with a 95\% confidence interval over five trials at different few-shot settings.}
\label{table: fewshot}
\begin{adjustbox}{max width=\textwidth}
\begin{tblr}{
  cells = {c},
  cell{1}{1} = {r=3}{},
  cell{1}{2} = {c=12}{},
  cell{2}{2} = {c=4}{},
  cell{2}{6} = {c=4}{},
  cell{2}{10} = {c=4}{},
  hline{1,13} = {-}{0.3em},
  hline{2,3,4} = {-}{0.1em},
  vline{2} = {-}{0.1em},
  vline{6, 10} = {-}{0.1em},
}
Models             & mIoU (\%) &            &             &            &             &            &            &            &             &             &             &             \\
                  & German       &            &             &            & T31TFM-1618 &            &            &            & PASTIS      &             &             &             \\
                  & 50-shot      & 20-shot    & 10-shot     & 5-shot     & 50-shot     & 20-shot    & 10-shot    & 5-shot     & 50-shot     & 20-shot     & 10-shot     & 5-shot      \\
MLP & $42.23_{\pm8.91}$ & $26.91_{\pm6.39}$ & $18.93_{\pm6.31}$ & $4.67_{\pm4.89}$ & $24.08_{\pm1.77}$ & $15.31_{\pm4.50}$ & $7.22_{\pm3.50}$ & $2.94_{\pm1.31}$ & $34.45_{\pm15.21}$ & $10.15_{\pm11.34}$ & $2.51_{\pm2.96}$ & $3.17_{\pm2.94}$ \\
DLinear & $27.25_{\pm1.14}$ & $39.45_{\pm0.52}$ & $27.55_{\pm20.64}$ & $9.24_{\pm3.06}$ & $16.92_{\pm0.60}$ & $9.57_{\pm0.62}$ & $5.59_{\pm2.03}$ & $4.41_{\pm1.81}$ & $47.39_{\pm3.78}$ & $49.23_{\pm23.63}$ & $21.57_{\pm5.29}$ & $23.97_{\pm27.15}$ \\
LSTM & $52.91_{\pm7.85}$ & $41.08_{\pm1.46}$ & $29.62_{\pm5.69}$ & $19.78_{\pm2.40}$ & $27.00_{\pm1.86}$ & $21.50_{\pm3.77}$ & $17.29_{\pm2.98}$ & $12.57_{\pm2.36}$ & $70.14_{\pm2.15}$ & $54.97_{\pm5.35}$ & $45.75_{\pm8.28}$ & $34.50_{\pm9.32}$ \\
DCM & $46.12_{\pm3.36}$ & $23.36_{\pm1.99}$ & $12.46_{\pm4.91}$ & $9.21_{\pm0.97}$ & $23.95_{\pm2.14}$ & $17.61_{\pm0.82}$ & $9.77_{\pm2.60}$ & $4.84_{\pm3.44}$ & $59.96_{\pm4.14}$ & $39.32_{\pm3.40}$ & $36.78_{\pm2.81}$ & $16.82_{\pm4.84}$ \\
TempCNN & $70.66_{\pm3.16}$ & $56.43_{\pm4.05}$ & $43.72_{\pm5.99}$ & $31.40_{\pm2.08}$ & $40.48_{\pm2.41}$ & $30.73_{\pm3.03}$ & $25.02_{\pm2.55}$ & $21.02_{\pm2.83}$ & $75.59_{\pm1.95}$ & $57.51_{\pm4.15}$ & $42.53_{\pm1.49}$ & $29.18_{\pm3.27}$ \\
TimesNet & $52.99_{\pm4.73}$ & $44.37_{\pm2.53}$ & $57.11_{\pm19.53}$ & $19.42_{\pm2.79}$ & $33.12_{\pm2.13}$ & $26.92_{\pm2.35}$ & $21.02_{\pm2.50}$ & $16.34_{\pm2.12}$ & $71.82_{\pm3.41}$ & $61.57_{\pm6.87}$ & $74.68_{\pm18.21}$ & $34.49_{\pm5.17}$ \\
Transformer & $54.14_{\pm7.82}$ & $39.34_{\pm3.09}$ & $23.83_{\pm2.59}$ & $16.69_{\pm1.58}$ & $29.52_{\pm1.18}$ & $24.22_{\pm2.31}$ & $19.51_{\pm1.67}$ & $14.13_{\pm2.04}$ & $74.34_{\pm3.49}$ & $58.70_{\pm4.47}$ & $41.10_{\pm4.72}$ & $28.74_{\pm3.93}$ \\
Informer & $54.03_{\pm2.08}$ & $39.68_{\pm4.10}$ & $24.40_{\pm3.10}$ & $17.60_{\pm0.76}$ & $30.51_{\pm1.06}$ & $24.77_{\pm2.45}$ & $19.24_{\pm1.28}$ & $12.87_{\pm1.80}$ & $72.40_{\pm3.37}$ & $57.80_{\pm3.25}$ & $42.70_{\pm4.59}$ & $28.33_{\pm2.65}$ \\
Ours & $50.25_{\pm2.83}$ & $39.47_{\pm2.34}$ & $33.78_{\pm3.77}$ & $28.52_{\pm2.19}$ & $32.26_{\pm0.68}$ & $25.36_{\pm1.87}$ & $20.37_{\pm1.09}$ & $16.98_{\pm1.08}$ & $62.63_{\pm2.21}$ & $49.31_{\pm2.88}$ & $41.13_{\pm2.06}$ & $33.93_{\pm3.57}$ \\
\end{tblr}
\end{adjustbox}
\end{table}

As shown in Figure~\ref{fig:confusionM}, the proposed method outperforms more than half of the deep learning models. Specifically, when \(n\) is set to 50, it significantly outperforms RNN-based and MLP-based models, while its performance is slightly lower than other SOTA models such as TimesNet and Informer. As \(n\) decreases to 20, it outperforms the Transformer by 0.13\% in mIoU on the German dataset. However, it lags slightly behind other Transformer-based models. As \(n\) decreases further to 10, it widens its performance gap with RNN-based and MLP-based models, although it trails CNN-based models in mIoU by an average of 24.51\% over the three datasets. This is consistent with the results of the Vision Transformer~\cite{dosovitskiy2020image}, which shows that the CNN-based models perform better in the few-shot setting. It is worth noting that when \(n\) falls to a minimum value of 5, our method outperforms almost all deep learning models on the three datasets. It is on average just under 3\% lower than TempCNN or TimesNet. As the value of \(n\) gradually decreases, the advantage of our method becomes more pronounced. This may be because compressors are data type agnostic and non-parametric methods have no underlying assumptions~\cite{jiang2023low}.

\section{Analyses}
Due to the low compression speeds resulting from pure numerical mapping and the inherently low compression speeds of the \(bz2\) compressor, we randomly sample 20\% of the total dataset for all experiments in this section. Half of this subset is divided as the training set and the other half as the test set. To ensure the generalisability of the experiment, we repeat it five times, randomly selecting a different subset each time and then calculating the average of the five results. The five random seeds are 2024, 21, 32, 400, 47.

\subsection{Different Alphabet Lengths}
\begin{figure}[htbp]
    \centering
    \subfloat{}{
        \includegraphics[width=0.5\textwidth]{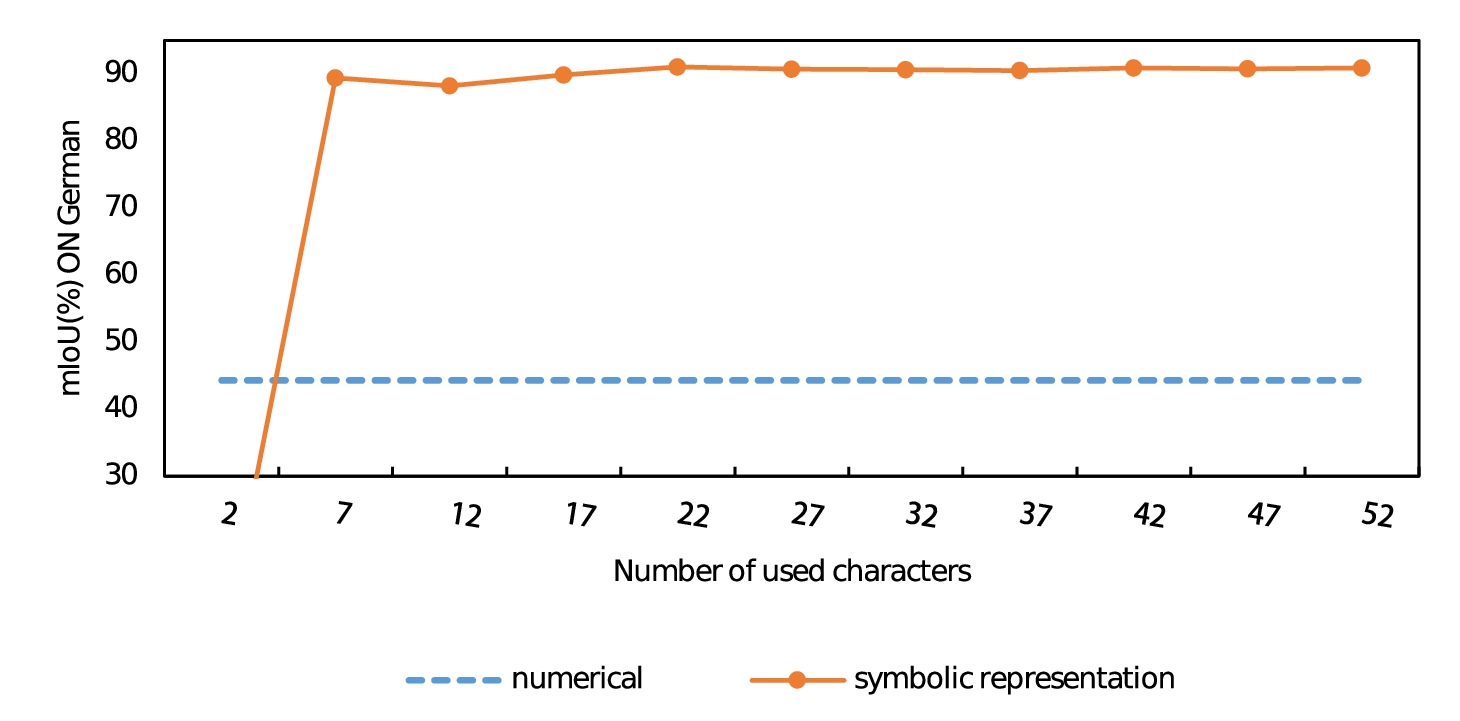}
        \label{fig:sub1}
    }
    \subfloat{}{
        \includegraphics[width=0.5\textwidth]{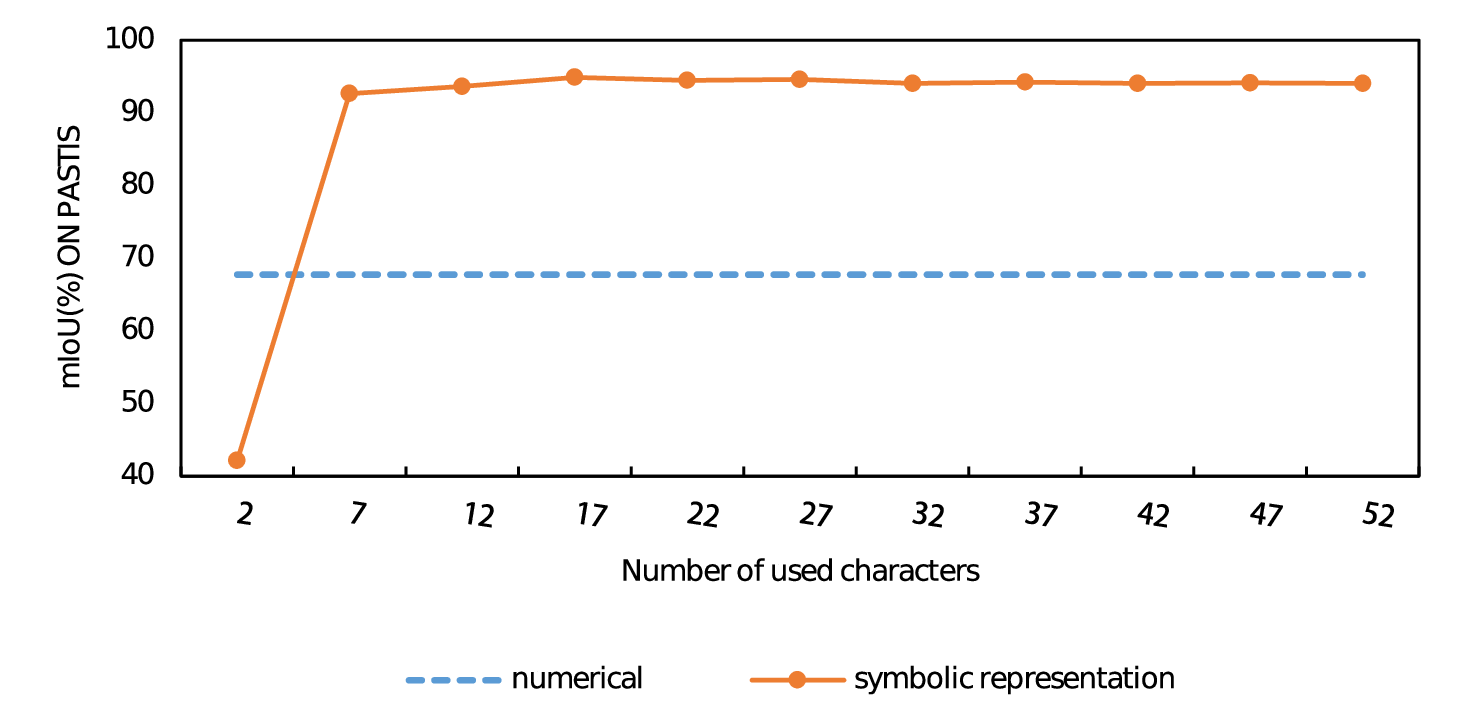}
        \label{fig:sub2}
    }
    \subfloat{}{
        \includegraphics[width=0.5\textwidth]{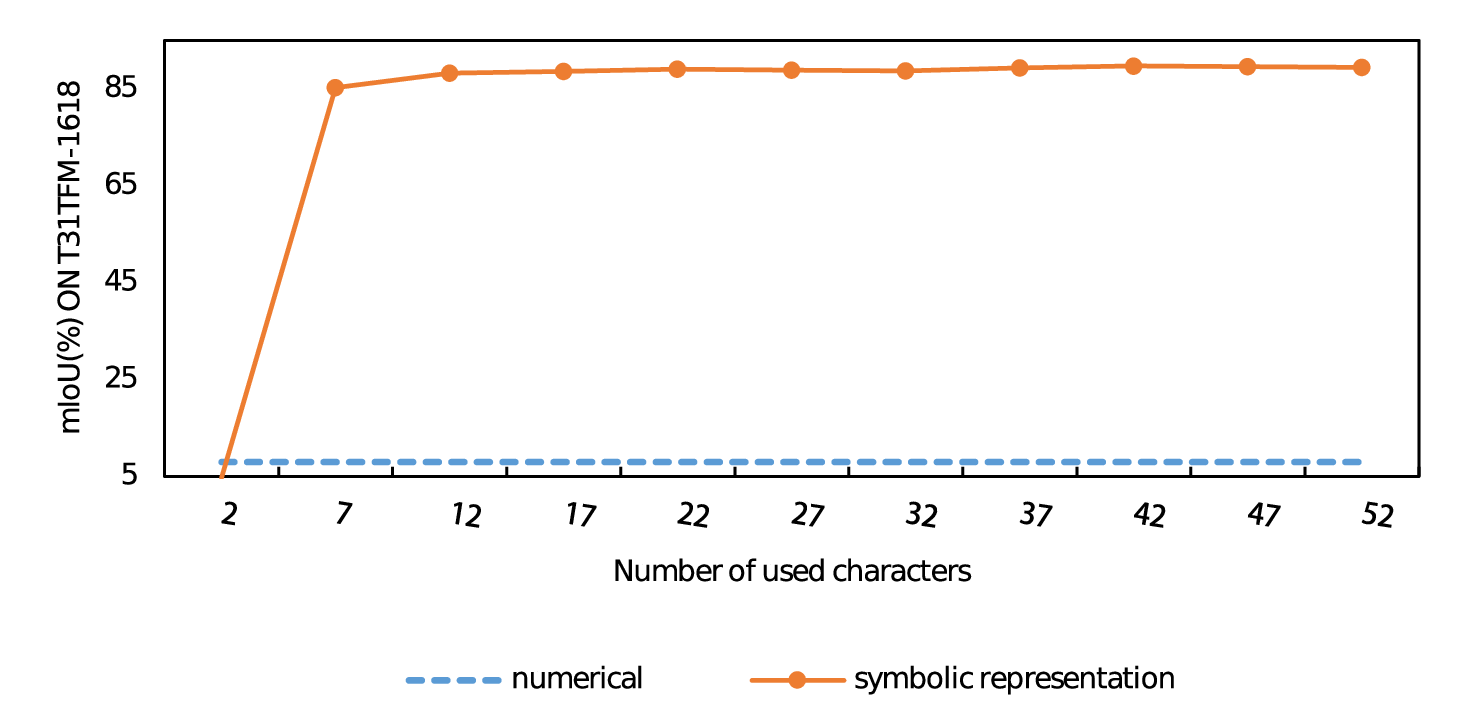}
        \label{fig:sub3}
    }

    \caption{Results on three datasets for different alphabet lengths.}
    \label{fig:Symbolic Representation}
\end{figure}

In the experiment, classification is performed by one group using pure numerical mapping, while the other group utilizes symbolic representation. The alphabet length starts at 2 and increases in increments of 5 up to 52 (including 26 lower case and 26 upper case letters). Figure~\ref{fig:Symbolic Representation} illustrates the effect of different mappings and different alphabet lengths on the classification results.

Compared to the symbolic representation mapping, the purely numerical mapping results in a worse classification performance. The mIoU is approximately 46\%, 80.71\% and 26.34\% lower on the three datasets respectively (excluding the case where the length of the alphabet \( l \) is equal to 2). One possible reason for this is that the compressor \textit{gzip}, which is widely used in the text domain~\cite{jiang2023lowresource}, is more adept at capturing patterns in symbolic representations. However, purely numerical information is not the same as character patterns.

\begin{figure}[htbp]
    \centering
    \includegraphics[width=\textwidth]{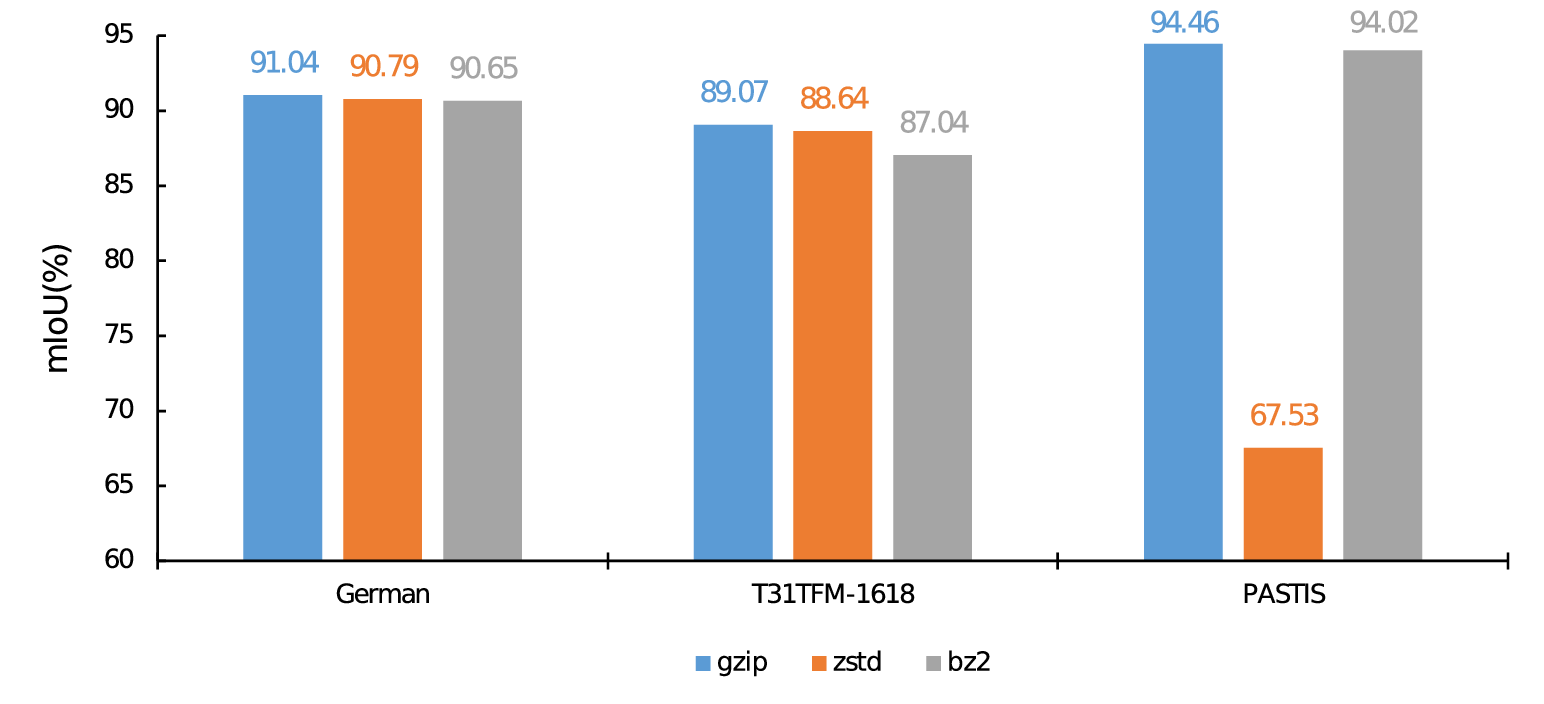}
    \caption{Results on three datasets for different compressors. `zstd' denotes the \textit{zstandard} compressor.}
    \label{fig:enter-label}
\end{figure}

When symbolic representation mappings are used, mIoU initially shows an upward trend. A possible reason for this is that rich features are lost when the variety of letters is too small. When the alphabet length exceeds 7, mIoU fluctuates within a range of no more than 3\%, indicating relatively stable performance overall. This also demonstrates the robustness of the proposed mapping method.

\subsection{Different Compressors}
The effect of different compressors on the classification results is shown in Figure~\ref{fig:enter-label}. On all three datasets, the classification performance of \textit{gzip} is significantly better than all other compressors, with an average mIoU of 91.52$\%$. This is consistent with the results of related research~\cite{jiang2023lowresource}. In comparison, the \textit{bz2}'s mIoUs are on average 0.95\% lower. The passable performance is probably due to the high compression ratio~\cite{jiang2023lowresource}. It is worth noting that on the PASTIS dataset, the classification accuracy of \textit{zstandard} is 25\% lower than that of other compressors, possibly due to the loss of fine-grained detail and changes in statistical properties during the compression process, which affect the classifier's performance.

\section{Conclusions}
In this research, we introduce a non-training alternative to deep learning models and bring compressor-based classification from  text classification to multi-spectral temporal crop classification. A novel Symbolic Representation Module is proposed to convert the reflectivities of all pixels into symbolic representations. These symbolic representations are then cross-transformed in both the channel and time dimensions to generate symbolic embeddings. Finally, based on the Multi-scale NCDs we have designed, crop classification is implemented using only a \(k\)NN. The results show that even without training, the proposed method achieves results comparable to those of large-scale trained deep learning models. It outperforms 5, 8, 5 advanced deep learning models on three benchmark datasets. It also outperforms more than half of these models in the few-shot setting with sparse labels. This lightweight and generalisation advantage contributes to its use in real-world agricultural production.

\subsubsection{Acknowledgements} 
We acknowledge Eren Yeager and Mikasa Ackerman for advising on the design of the model. This work was supported by Mathematics Research Branch Institute of Beijing Association of Higher Education \& Beijing Interdisciplinary Science Society.

%
%
%
\bibliographystyle{splncs04}
\bibliography{cas-refs}

\end{document}